# A Model Approximation Scheme for Planning in Partially Observable Stochastic Domains

**Nevin L. Zhang**                                      LZHANG@CS.UST.HK
**Wenju Liu**                                           WLIU@CS.UST.HK
*Department of Computer Science*
*Hong Kong University of Science and Technology*
*Hong Kong, China*

## Abstract

Partially observable Markov decision processes (POMDPs) are a natural model for planning problems where effects of actions are nondeterministic and the state of the world is not completely observable. It is difficult to solve POMDPs exactly. This paper proposes a new approximation scheme. The basic idea is to transform a POMDP into another one where additional information is provided by an oracle. The oracle informs the planning agent that the current state of the world is in a certain region. The transformed POMDP is consequently said to be region observable. It is easier to solve than the original POMDP. We propose to solve the transformed POMDP and use its optimal policy to construct an approximate policy for the original POMDP. By controlling the amount of additional information that the oracle provides, it is possible to find a proper tradeoff between computational time and approximation quality. In terms of algorithmic contributions, we study in details how to exploit region observability in solving the transformed POMDP. To facilitate the study, we also propose a new exact algorithm for general POMDPs. The algorithm is conceptually simple and yet is significantly more efficient than all previous exact algorithms.

## 1. Introduction

In a completely observable and deterministic world, to plan is to find a sequence of actions that will lead an agent to achieve a goal. In real-world applications, the world is rarely completely observable and effects of actions are almost always nondeterministic. For this reason, a growing number of researchers concern themselves with planning in partially observable stochastic domains (e.g., Dean & Wellman, 1991; Cassandra *et al.*, 1994; Parr & Russell, 1995; Boutilier & Poole, 1996). Partially observable Markov decision processes (POMDPs) can be used as a model for planning in such domains. In this model, nondeterminism in effects of actions is encoded by transition probabilities, partial observability of the world by observation probabilities, and goals and criteria for good plans by reward functions (see Section 2 for details).

POMDPs are classified into *finite horizon* POMDPs and *infinite horizon* POMDPs depending on the number of time points considered. Infinite horizon POMDPs are usually used for planning because one typically does not know beforehand the number of steps it takes to achieve a goal. This paper is concerned with how to solve an infinite horizon POMDP.

## 1.1 Difficulties in Solving POMDPs

When the world is fully observable, a POMDP reduces to a *Markov decision process* (MDP). MDPs have been studied extensively in the dynamic-programming literature (e.g., Puterman, 1990; Bertsekas, 1987, White, 1993). Recent works have concentrated on how to deal with large state spaces (Dean *et al.*, 1993; Boutilier *et al.*, 1995; Dean & Lin, 1995).

We are concerned with the partially observable case. This case is considerably more difficult than the fully observable case for two related reasons. First, when the agent knows exactly in which state the world currently is, information from the past (past observations and actions) is irrelevant to the current decision. This is the Markov property. On the other hand, when the agent does not fully observe the state of the world, past information becomes relevant because it can help the agent to better estimate the current state of the world. The problem is that the number of possible states of past information increases exponentially with time.

Second, in MDPs the effects of an action are fully observed at the next time point. In POMDPs, on the other hand, the effects of an action are not fully observed at the next time point. Hence one cannot clearly tell the effects of the current action from those of the agent's future behaviors. To properly evaluate the effects of an action, one needs to look into the future and consider the combination of the action with each of the agent's possible behaviors in a, possibly large, number of future steps. The problem is that the number of ways the agent can behave is exponential in the number of future steps considered.

## 1.2 Previous Work

Previous methods for solving POMDPs are usually classified into exact methods and approximate methods (Lovejoy, 1991a). They can also can be classified according to which of the aforementioned two difficulties they directly address. Most previous methods address the difficulty of exponential number of future behaviors (Sondik, 1971; Sondik & Mendelssohn, 1979; Monahan, 1982; Cheng, 1988; Lovejoy, 1991b; and Cassandra *et al.*, 1994). They prune from consideration behaviors that can never be optimal no matter what the information state is (Section 4). Other methods deal with the problem of exponential number of past information states either by aggregating them (Platzman, 1977; White & Schere, 1994) or by considering only a subset of them (Lovejoy, 1992; Brafman, 1997; Hauskrecht, 1997). They are approximation methods in nature.

## 1.3 Model Approximation

In previous approximation methods, approximation takes place in the process of solving a POMDP. We advocate *model approximation methods*. Such a method approximates a POMDP itself by another one that is easier to solve and uses the solution of the latter to construct an approximate solution to the original POMDP.

Model approximation can be in the form of a more informative observation model, or a more deterministic action model, or a simpler state space, or a combination of two or all of the three alternatives. Cassandra *et al.* (1996) proposed to approximate POMDPs by using MDPs. This is an example of model approximation in the form of a more informative observation model. There is also some work on reducing the size of the state spaces of MDPs by aggregation (e.g., Bertsekas & Castanon, 1989; Dean & Lin, 1995; Dean & Givan, 1997). Such work can be conceivably extended to POMDPs, leading to model approximation in the form of a simpler state space. We are not aware of any model approximation schemes in the form of a more deterministic action model.

## 1.4 Our Proposal

This paper proposes a new model approximation scheme in the form of a more informative observation model. It is a generalization of the idea of approximating POMDPs by using MDPs.

We transform a POMDP by assuming that, in addition to the observations obtained by itself, the agent also receives a report from an oracle who knows the true state of the world. The oracle does not report the true state itself. Instead, it selects, from a predetermined list of candidate regions, a region that contains the true state and reports that region. The transformed POMDP is said to be *region observable* because the agent knows for sure that the true state is in the region reported by the oracle.

When all candidate regions are singletons, the oracle actually reports the true state of the world. The region observable POMDP reduces to an MDP. MDPs are much easier to solve than POMDPs. One would expect the region observable POMDP to be solvable when all candidate regions are small.

In terms of approximation quality, the larger the candidate regions, the less additional information the oracle provides and hence the more accurate the approximation. In the extreme case when there is only one candidate region and it consists of all possible states of the world, the oracle provides no additional information at all. Consequently, the region observable POMDP is identical to the original POMDP.

A method for determining approximation quality will be described later in this paper. It allows one to make the tradeoff between approximation quality and computational time as follows: start with small candidate regions and increase their sizes gradually until the approximation becomes accurate enough or the region observable POMDP becomes untractable.

Due to problem characteristics, accurate approximation can usually be achieved with small candidate regions. In many applications the agent often has a good idea about the state of the world (e.g., Simmons & Koenig, 1995). Take robot path planning as an example. Observing a landmark, a room number for instance, would imply that the robot is at the proximity of that landmark. Observing a feature about the world, a corridor T-junction for instance, might imply the robot is in one of several regions. Taking history into account, the robot might be able to determine a unique region for its current location. Also, an action usually moves the state of the world to only a few "nearby" states. Thus if the robot has a good idea about the current state of world, it should continue to have a good idea about it in the next few steps.

When the agent has a good idea about the state of the world at all times, the oracle does not provide much additional information even with small candidate regions and hence approximation is accurate. Region observable POMDPs with small candidate regions are much easier to solve than general POMDPs.

## 1.5 Organization

We will first show how POMDPs can be used as a model for planning in partially observable stochastic domains (Section 2) and give a concise review of the theory of POMDPs (Sections 3 and 4). We will then propose a new method for dynamic-programming updates, a key step in algorithms that solve POMDPs via value iteration (Section 5). Thereafter, we will formally introduce the concept of region observable POMDPs (Section 6) and develop an algorithm for solving region observable POMDPs (Sections 7, 8, and 9). In Section 10, we will discuss decision making for the original POMDPs based on the solutions of their region observable approximations, followed by a method for determining approximation quality (Section 11) and a method to make the tradeoff between approximation quality and computational time (Section 12). Finally, empirical results will be reported in Section 13 and conclusions will be provided in Section 14.

## 2. Planning in Stochastic Domains and POMDPs

To specify a planning problem, one needs to give a set $\mathcal{S}$ of possible states of the world, a set $\mathcal{O}$ of possible observations, and a set $\mathcal{A}$ of possible actions. The sets $\mathcal{O}$ and $\mathcal{A}$ are always assumed to be finite in the literature, while the state space $\mathcal{S}$ can be continuous as well as finite. In this paper, we consider only finite state space. One needs also to give an observation model which describes the relationship between observations and the state of the world, and an action model which describes effects of actions.

As a background example, consider path planning for a robot who acts in an office environment. Here $\mathcal{S}$ is the set of all location-orientation pairs, $\mathcal{O}$ is the set of possible sensor readings, and $\mathcal{A}$ consists of actions `move-forward`, `turn-left`, `turn-right`, and `declare-goal`.

The current observation $o$ depends on the current state of the world $s$. Due to sensor noise, this dependency is uncertain in nature. The observation $o$ sometimes also depends on the action that the robot has just taken $a_-$. The minus sign in the subscript indicates the previous time point. In the POMDP model, the dependency of $o$ upon $s$ and $a_-$ is numerically characterized by a conditional probability $P(o|s, a_-)$, which is usually referred to as the *observation probability*. It is the observation model.

In a region observable POMDP, the current observation also depends on the previous state of the world $s_-$. The observation probability for this case can be written as $P(o|s, a_-, s_-)$.

The state $s_+$ the world will be in at the next time point depends on the current action $a$ and the current state $s$. The plus sign in the subscript indicates the next time point. This dependency is again uncertain in nature due to uncertainty in the actuator. In the POMDP model, the dependency of $s_+$ upon $s$ and $a$ is numerically characterized by a conditional probability $P(s_+|s, a)$, which is usually referred to as the *transition probability*. It is the action model.

On many occasions, we need to consider the joint conditional probability $P(s_+, o_+|s, a)$ of the next state of the world and the next observation given the current state and the current action. It is given by

$$P(s_+, o_+|s, a) = P(s_+|s, a)P(o_+|s_+, a, s).$$

Knowledge about the initial state, if available, is represented as a probability distribution $P_0$ over $\mathcal{S}$. When the agent knows the initial state with certainty, $P_0$ is 1 at the initial state and 0 everywhere else. The planning goal is encoded by a *reward function* such as the following:

$$r(s, a) = \left\{ \begin{array}{ll} 1 & \text{if } a=\texttt{declare-goal} \text{ and } s=\texttt{goal}, \\ 0 & \text{otherwise}. \end{array} \right.$$

The preference for short plans is encoded by discounting future rewards with respect to the current reward (see the next section).

In summary, a POMDP consists of a set of possible states of the world, a set of possible observations, a set of possible actions, a observation probability, a transition probability, and a reward function. An MDP has the same ingredients as an POMDP except that it has no observation probability. This is because the state of the world is completely observed in an MDP.

## 3. Basics of POMDPs

This section reviews several concepts and results related to POMDPs.

### 3.1 Belief States

In a POMDP, an agent chooses and executes an action at each time point. The choice is made based on information from the past (past observations and past actions) and the current observation. The amount of memory required to store past observations and actions increases linearly with time. This makes it difficult to maintain past information after a long period of time.

The standard way to overcome this difficulty is to maintain, instead of past information, the agent's *belief state* — the probability distribution $P(s_t|o_t, a_{t-1}, o_{t-1}, \ldots, a_1, o_1, P_0)$ of the current state $s_t$ of the world given past information and the current observation. It is well known that the belief state is a *sufficient statistic* in the sense that it captures all the information contained in past information and the current observation that is useful for action selection. Hence the agent can base its decision solely on the belief state.

Compared with maintaining past information, maintaining the belief state is desirable because the number of possible states of the world is finite. One needs only to maintain a fixed and finite number of probability values[1].

The initial belief state is $P_0$. One question is how the agent should update its belief state as time goes by. Following Littman (1994), we use $b$ to denote a belief state. For any state $s$, $b(s)$ is the probability that the world is in state $s$. The set of all possible belief states will be denoted by $\mathcal{B}$.

Suppose $b$ is the current belief state, and $a$ is the current action. If the observation $o_+$ is obtained at the next time point, then the agent should update its belief state from $b$ to a new belief state $b_+$ given by

$$b_+(s_+) = k \sum_s P(s_+, o_+|s, a)b(s), \tag{1}$$

---

1. Storing these values exactly could require unbounded precision. Approximations are implicitly being made due to the fact that machine precision is bounded.

where $k=1/\sum_{s,s_+} P(s_+, o_+|s, a)b(s)$ is the normalization constant (e.g., Littman, 1994).

## 3.2 POMDPs as MDPs

For any belief state $b$ and any action $a$, define

$$r(b, a) = \sum_s b(s)r(s, a). \tag{2}$$

It is the expected immediate reward for taking action $a$ in belief state $b$.

For any belief state $b$, any action $a$, and any observation $o_+$, define

$$P(o_+|b, a) = \sum_{s,s_+} P(s_+, o_+|s, a)b(s). \tag{3}$$

It is the probability of observing $o_+$ at the next time point given that the current belief state is $b$ and the current action is $a$. Let $b_+$ be the belief state given by equation (1). $P(o_+|b, a)$ can also be understood as the probability of the next belief state being $b_+$ given that the current action is $a$ and the current belief state is $b$.

A POMDP over world state space $\mathcal{S}$ can be viewed as an MDP over the belief state space $\mathcal{B}$. The the reward function and the transition probability of the MDP are given by Equations (2) and (3) respectively.

## 3.3 Optimal Policies

At each time point, the agent consults its belief state and chooses an action. A *policy* $\pi$ prescribes an action for each possible belief state. Formally it is a mapping from $\mathcal{B}$ to $\mathcal{A}$. For each belief state $b$, $\pi(b)$ is the action prescribed by $\pi$ for $b$.

Suppose $b_0$ is the current belief state. If an agent follows a policy $\pi$, then its current action is $\pi(b_0)$ and the immediate reward is $r_0(b, \pi(b_0))$; with probability $P(o_+|b_0, \pi(b_0))$, the agent's next belief state $b_1$ will be as given by Equation (1), the next action will be $\pi(b_1)$, and the next reward will be $r_1(b_1, \pi(b_1))$; and so on and so forth. The quality of a policy is measured by the expected discounted rewards it garners. Formally the *value function* of a policy $\pi$ is defined for each belief state $b_0$ to be the following expectation:

$$V^\pi(b_0) = E_{b_0}\left[\sum_{i=0}^\infty \gamma^i r_i(b_i, \pi(b_i))\right], \tag{4}$$

where $0 \leq \gamma < 1$ is the *discount factor*.

A policy $\pi_1$ *dominates* another policy $\pi_2$ if for each belief state $b \in \mathcal{B}$

$$V^{\pi_1}(b) \geq V^{\pi_2}(b). \tag{5}$$

Domination is a partial ordering among policies. It is well known that there exists a policy that dominates all other policies (e.g., Puterman, 1990). Such a policy is called an *optimal policy*. The value function of an optimal policy is called the *optimal value function* and is denoted by $V^*$.

### 3.4 Value Iteration

Value iteration is a standard way for solving infinite horizon MDPs (Bellman, 1957). It begins with an arbitrary initial function $V_0^*(b)$ and improves it iteratively by using the following equation

$$V_t^*(b) = max_a[r(b, a) + \gamma \sum_{o_+} P(o_+|b, a)V_{t-1}^*(b_+)], \tag{6}$$

where $b_+$ is the belief state given by equation (1). If $V_0^*=0$, then $V_t^*$ is called the *t-step optimal value function*. For any belief state $b$, $V_t^*(b)$ is the optimal expected reward the agent can get in $t$ steps starting from $b$.

The following theorem (Puterman, 1990, page 361) tells one when to terminate value iteration and how to construct a "good enough" policy.

**Theorem 1** *Let $\pi$ the policy given by*

$$\pi(b) = arg \ max_a[r(b, a) + \gamma \sum_{o_+} P(o_+|b, a)V_t^*(b_+)]. \tag{7}$$

*If $max_{b \in \mathcal{B}}|V_t^*(b) - V_{t-1}^*(b)| \leq \epsilon$, then*

$$max_{b \in \mathcal{B}}|V^\pi(b) - V^*(b)| \leq \frac{2\epsilon\gamma}{1 - \gamma}. \square \tag{8}$$

The quantity $max_{b \in \mathcal{B}}|V_t^*(b) - V_{t-1}^*(b)|$ is sometimes called the *Bellman residual* and the policy $\pi$ is called the *greedy policy* based on $V_t^*$.

Algorithms for POMDPs are classified into *exact* or *approximation* algorithms depending on whether they compute the *t*-step optimal value function $V_t^*$ exactly (Lovejoy, 1991a). In the next two sections, we discuss the theoretical foundations of exact algorithms and develop a new exact algorithm. Thereafter, we propose a new approximation algorithm.

## 4. Piecewise Linearity and Implicit Value Iteration

Since the belief space is continuous, exact value iteration cannot be carried out explicitly. Fortunately, it can be carried out implicitly due to the piecewise linearity of the *t*-step optimal value functions. To explain piecewise linearity, we need the concept of policy trees.

### 4.1 Policy Trees

A *t-step policy tree* $p_t$ (Littman, 1994) prescribes an action for the current time point and an action for each possible information scenario $(o_1, \ldots, o_i, a_0, \ldots, a_{i-1})$ at each of the next $t-1$ time points $i$. Figure 1 shows a 3-step policy tree. The tree reads as follows. `Move-forward` at the current time point. At the next time point, if $o_1=0$ is observed then `turn-left`. Thereafter if $o_2=0$ is observed then `turn-left` again; else if $o_2=1$ is observed then `declare-goal`; else if $o_2=2$ is observed then `move-forward`. And so on and so forth. To relate back to the introduction, a *t*-step policy tree prescribes a way the agent might behave at the current and the next $t-1$ time points.

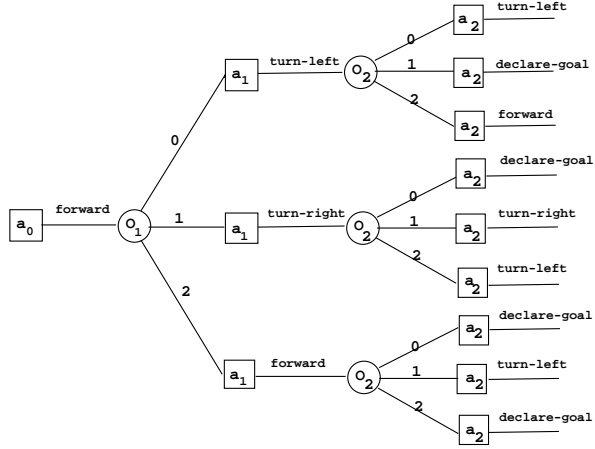

Figure 1: A 3-step policy tree.

When $t>1$, the subtree rooted at the $o_1$ node will be called a *o-rooted $t-1$-step policy tree*, and will be denoted by $\delta^{t-1}$. It is a mapping from the set of possible observations $\mathcal{O}$ to the set of all possible $t-1$-step policy trees; it prescribes a $t-1$ step policy tree $\delta^{t-1}(o)$ for each possible observation $o$. In our example, $\delta^2(o_1=0)$ is the 2-step policy tree rooted at the uppermost $a_1$ node.

When $t>1$, a $t$-step policy tree $p_t$ has two components: an action $a$ for the current time point and an $o$-rooted $t-1$-step policy tree $\delta^{t-1}$ for the next $t-1$ time points. For this reason, we shall sometimes write $p_t$ as a pair $(a, \delta^{t-1})$ and call $a$ the *first action* of $p_t$.

By altering the actions on the edges out of the $a$-nodes, one obtains different $t$-step policy trees. The set of all possible $t$-step policy trees will be denoted by $\mathcal{P}_t$. A 1-step policy tree is simply an action, and hence $\mathcal{P}_1$ is the same as the set of possible actions $\mathcal{A}$.

### 4.2 State Value Functions of Policy Trees

For any state $s$ and any $t$-step policy tree $p_t=(a, \delta^{t-1})$, recursively define

$$V_{p_t}(s) = r(s, a) + \gamma \sum_{o_+} \sum_{s_+} V_{\delta^{t-1}(o_+)}(s_+) P(s_+, o_+ | s, a), \tag{9}$$

where the second term is to be understood as 0 when $t=1$. It is the expected discounted total reward the agent receives at the current time and during the next $t-1$ time points if the world is currently in state $s$ and the agent behaves according to the policy tree $p_t$. We call $V_{p_t}$ the *state value function* of the $t$-step policy tree $p_t$.

Without mentioning the policy tree, we shall sometimes call $V_{p_t}$ a *$t$-step state value function*. The collection of all $t$-step state value functions will be denoted by $\mathcal{V}_t$, i.e.

$$\mathcal{V}_t = \{V_{p_t} | p_t \in \mathcal{P}_t\}.$$

For convenience, we let $\mathcal{V}_0$ consist of one single function of $s$ that is zero for all $s$.

### 4.3 State Space Functions and Belief Space Functions

It is worthwhile to point out that a $t$-step state value function is a *state space function*, i.e. a function over the state space $\mathcal{S}$, while the $t$-step optimal value function is a *belief space function*, i.e. a function over the belief space $\mathcal{B}$. We often use notations such as $\alpha$ or $\beta$ to refer to state space functions. A state space function $\alpha(s)$ *induces* a belief space function through

$$\alpha(b) = \sum_s \alpha(s)b(s).$$

Regarding $b$ as a vector with one component $b(s)$ for each $s$, the induced belief space function is a *linear* combination of the components of $b$. For convenience, we simply say that $\alpha(b)$ is *linear* in $b$.

A collection $\mathcal{V}$ of state space functions *induces* a belief space function through

$$\mathcal{V}(b) = max_{\alpha \in \mathcal{V}} \alpha(b). \tag{10}$$

Note that we are using $\mathcal{V}$ to denote both a set of state space functions and the belief space function it induces. When $\mathcal{V}$ is empty, $\mathcal{V}(b)$ is 0 by definition.

The induced belief space function $\mathcal{V}(b)$ is *piecewise linear* in $b$ in the sense that, for each $\alpha \in \mathcal{V}$, it equals $\alpha(b)$ in the region $\{b|\alpha(b) \geq \beta(b)$ for any other $\beta \in \mathcal{V}\}$ of the belief space $\mathcal{B}$ and hence is linear in $b$ in that region.

### 4.4 Piecewise Linearity of Optimal Value Functions

The following theorem was first proved by Sondik (1971). It first appeared in its present form in Littman (1994).

**Theorem 2** *(Piecewise Linearity) The $t$-step optimal value function $V_t^*$ is the same as the belief space function induced by the collection of all $t$-step state value functions $\mathcal{V}_t$, i.e. for any belief state $b$*

$$V_t^*(b) = \mathcal{V}_t(b). \square$$

The theorem is true for the following reasons. $V_t^*(b)$ is the reward the agent receives if it behaves optimally and for any policy tree $p_t$, $V_{p_t}(b)$ is the reward the agent gets if it behaves according to $p_t$. Because one of the policy trees must be optimal, $V_t^*(b) = max_{p_t} V_{p_t}(b) = \mathcal{V}_t(b)$. Due to this theorem, we say that the collection $\mathcal{V}_t$ of state value functions is *representation* of $V_t^*$.

### 4.5 Parsimonious Representations

The size of $\mathcal{V}_t$ increases exponentially with $t$. As a matter of fact, the total number of $t$-step policy trees (Cassandra, 1994) is:

$$|\mathcal{P}_t| = |\mathcal{A}|^{\frac{|\mathcal{O}|^t-1}{|\mathcal{O}|-1}}.$$

There are potentially the same number of $t$-step state value functions. Fortunately, many of the state value functions can be pruned without affecting the induced belief space function. Let us make this property more explicit.

Let $\mathcal{W}$ and $\mathcal{X}$ be two sets of state space functions. We say that $\mathcal{W}$ *covers* $\mathcal{X}$ if it induces the same belief space function as $\mathcal{X}$ does, i.e. if

$$\mathcal{W}(b) = \mathcal{X}(b)$$

for any belief state $b$. We say that $\mathcal{W}$ *parsimoniously covers* $\mathcal{X}$ if $\mathcal{W}$ covers $\mathcal{X}$ and none of its proper subsets do. When $\mathcal{W}$ covers or parsimoniously covers $\mathcal{X}$, we refer to $\mathcal{W}$ as a *covering* or a *parsimonious covering* of $\mathcal{X}$.

**Theorem 3** *All parsimonious coverings of a set of state space functions consist of the same number of state space functions.* □

The theorem has been known for sometime (e.g., Littman, 1994). Due to this theorem, one can also define a parsimonious covering as a covering that contains the minimum number of state space functions.

A parsimonious covering $\hat{\mathcal{V}}_t$ of $\mathcal{V}_t$ is also a representation of $V_t^*$ in the sense that $\hat{\mathcal{V}}_t(b) = V_t^*(b)$ for any belief state $b$. This representation is parsimonious because it consists of the fewest number of state space functions among all the representations of $V_t^*$.

## 4.6 Dynamic-Programming Updates

The question now is how to obtain a parsimonious covering of $\mathcal{V}_t$. As will be shown in the next section, it is possible to obtain a parsimonious covering of $\mathcal{V}_t$ by starting from a parsimonious covering of $\mathcal{V}_{t-1}$. The process of computing a parsimonious covering of $\mathcal{V}_t$ from a parsimonious covering of $\mathcal{V}_{t-1}$ is called *dynamic-programming updates* (Littman *et al.*, 1995). It is a key step in algorithms that solve POMDPs via value iteration.

Previous algorithms for dynamic-programming updates include the enumeration and pruning algorithms by Monahan (1992), Eagle (1984), and Lark (White, 1991), the one-pass algorithm by Sondik (1971), the linear support and relaxed region algorithms by Cheng (1988), and the witness algorithm by Cassandra *et al.* (1994) and Littman (1994). The witness algorithm has been empirically proved to be the most efficient among all those algorithms (Littman *et al.*, 1995).

## 4.7 Implicit Value Iteration

The procedure `solvePOMDP` shown in Figure 2 carries out value iteration implicitly: instead inductively computing the $t$-step optimal value function $V_t^*$ itself, it computes a parsimonious covering of $\mathcal{V}_t$ — a set of state space functions that represents $V_t^*$. In the procedure, the subroutine `update`($\hat{\mathcal{V}}_{t-1}$) takes a parsimonious covering $\hat{\mathcal{V}}_{t-1}$ of $\mathcal{V}_{t-1}$ and returns a parsimonious covering $\hat{\mathcal{V}}_t$ of $\mathcal{V}_t$. It can be implemented using any of the algorithms mentioned in the previous subsection. The subroutine `stop`($\hat{\mathcal{V}}_t, \hat{\mathcal{V}}_{t-1}, \epsilon$) determines whether the Bellman residual has fallen below the threshold $\epsilon$ from the parsimonious coverings $\hat{\mathcal{V}}_{t-1}$ and $\hat{\mathcal{V}}_t$ of $\mathcal{V}_{t-1}$ and $\mathcal{V}_t$. See Littman (1994) for an implementation of this subroutine.

Procedure `solvePOMDP` terminates when the Bellam residual falls below the threshold $\epsilon$ and return a set of state space functions. The set $\hat{\mathcal{V}}_t$ of state space functions returned represents the $t$-step optimal value function $V_t^*$. It is the solution to the input POMDP. The planning agent keeps $\hat{\mathcal{V}}_t$ in its memory. When it needs to make a decision, the agent consults its belief state $b$ and chooses an action using (7) with $V_t^*(b_+)$ replaced by $\hat{\mathcal{V}}_t(b_+)$.

---

Procedure `solvePOMDP`($\mathcal{M}, \epsilon$):
- Input: $\mathcal{M}$ — A POMDP,
  $\epsilon$ — A positive number.
- Output: A set of state space functions.
1. $t \leftarrow 0$, $\hat{\mathcal{V}}_0 \leftarrow \{0\}$.
2. **Do**
   - $t = t + 1$.
   - $\hat{\mathcal{V}}_t \leftarrow$ `update`($\hat{\mathcal{V}}_{t-1}$).
   **while** `stop`($\hat{\mathcal{V}}_t, \hat{\mathcal{V}}_{t-1}, \epsilon$) = `no`.
3. Return $\hat{\mathcal{V}}_t$.

---

Figure 2: Implicit value iteration.

## 5. A New Algorithm for Dynamic-Programming Updates

This section proposes a new algorithm for dynamic-programming updates. There are four subsections. In the first three subsections, we show that a parsimonious covering of $\mathcal{V}_t$ can be obtained by starting from a parsimonious covering of $\mathcal{V}_{t-1}$ and, while doing so, introduce concepts and results that are necessary for the development of the new algorithm.

### 5.1 Relationship Between $\mathcal{V}_{t-1}$ and $\mathcal{V}_t$

Suppose $\mathcal{W}$ and $\mathcal{X}$ are two sets of state space functions. The *cross sum* $\mathcal{W} \bigoplus \mathcal{X}$ of $\mathcal{W}$ and $\mathcal{X}$ is the following set of state space functions:

$$\mathcal{W} \bigoplus \mathcal{X} = \{\alpha + \beta | \alpha \in \mathcal{W}, \beta \in \mathcal{X}\}.$$

It is evident that the cross sum operation is commutative and associative. Consequently, we can talk about the cross sum $\bigoplus_{i=0}^N \mathcal{W}_i$ of a list of sets $\mathcal{W}_0, \ldots, \mathcal{W}_N$ of state space functions.

For any action $a$ and any observation $o_+$, define

$$\mathcal{Q}_{a,o_+} = \{\gamma \sum_{s_+} \alpha(s_+) P(s_+, o_+ | s, a) | \alpha \in \mathcal{V}_{t-1}\}. \tag{11}$$

Note that since $a$ and $o_+$ are given, each member $\gamma \sum_{s_+} \alpha(s_+) P(s_+, o_+ | s, a)$ of the above set is a function of $s$, in other words, a state space function. Hence $\mathcal{Q}_{a,o_+}$ is a set of state space functions. Let $0, 1, \ldots, N$ be an enumeration of all possible values of $o_+$. We use $\bigoplus_{o_+} \mathcal{Q}_{a,o_+}$ to denote the cross sum $\bigoplus_{i=0}^N \mathcal{Q}_{a,i}$.

**Proposition 1** $\mathcal{V}_t = \cup_a[\{r(s,a)\} \bigoplus (\bigoplus_{o_+} \mathcal{Q}_{a,o_+})]$.

**Proof:** By the definition of the set $\mathcal{V}_t$ and Equation (9), a state space function $\alpha$ is in $\mathcal{V}_t$ if and only if there exist action $a$ and $\beta_{o_+} \in \mathcal{V}_{t-1}$ for each $o_+ \in \mathcal{O}$ such that

$$\alpha(s) = r(s, a) + \sum_{o_+} \gamma \sum_{s_+} \beta_{o_+}(s_+) P(s_+, o_+|s, a).$$

The proposition follows. $\square$

## 5.2 Properties of Coverings

**Lemma 1** *Suppose $\mathcal{W}$, $\mathcal{X}$, and $\mathcal{Y}$ are three sets of state space functions. If $\mathcal{W}$ parsimoniously covers $\mathcal{X}$ and $\mathcal{X}$ covers $\mathcal{Y}$, then $\mathcal{W}$ parsimoniously covers of $\mathcal{Y}$.* $\square$

**Lemma 2** *Let $\mathcal{W}$, $\mathcal{W}'$, $\mathcal{X}$, and $\mathcal{X}'$ be four sets of state space functions. If $\mathcal{W}'$ covers $\mathcal{W}$ and $\mathcal{X}'$ covers $\mathcal{X}$, then*

   *1. $\mathcal{W}' \bigoplus \mathcal{X}'$ covers $\mathcal{W} \bigoplus \mathcal{X}$.*

   *2. $\mathcal{W}' \cup \mathcal{X}'$ covers $\mathcal{W} \cup \mathcal{X}$.* $\square$

## 5.3 Coverings of $\mathcal{V}_t$ from Parsimonious Coverings of $\mathcal{V}_{t-1}$

Let $\hat{\mathcal{V}}_{t-1}$ be a parsimonious covering of $\mathcal{V}_{t-1}$. For any action $a$ and any observation $o_+$, define

$$\mathcal{Q}'_{a,o_+} = \{\gamma \sum_{s_+} \alpha(s_+) P(s_+, o_+|s, a)|\alpha \in \hat{\mathcal{V}}_{t-1}\}.$$

Note that the definition of $\mathcal{Q}'_{a,o_+}$ is the same as that of $\mathcal{Q}_{a,o_+}$ except that $\mathcal{V}_{t-1}$ is replaced by $\hat{\mathcal{V}}_{t-1}$. Also define

$$\mathcal{V}'_t = \cup_a [\{r(s, a)\} \bigoplus (\bigoplus_{o_+} \mathcal{Q}'_{a,o_+})].$$

**Proposition 2** *The set $\mathcal{V}'_t$ covers $\mathcal{V}_t$.*

Formal proof of this proposition is given in Appendix A. Informally, the fact that $\hat{\mathcal{V}}_{t-1}$ covers $\mathcal{V}_{t-1}$ implies that $\mathcal{Q}'_{a,o_+}$ covers $\mathcal{Q}_{a,o_+}$, which in turn implies that $\mathcal{V}'_t$ covers $\mathcal{V}_t$ due to Proposition 1 and Lemma 2.

According to Proposition 2 and Lemma 1, one can obtain a parsimonious covering of $\mathcal{V}_t$ by finding a parsimonious covering of $\mathcal{V}'_t$ and $\mathcal{V}'_t$ is defined in terms of a parsimonious covering $\hat{\mathcal{V}}_{t-1}$ of $\mathcal{V}_{t-1}$. This is why we said that a parsimonious covering of $\mathcal{V}_t$ can be obtained by starting from a parsimonious covering of $\mathcal{V}_{t-1}$.

Monahan's exhaustive method finds a parsimonious covering of $\mathcal{V}'_t$ by enumerating all the state space functions in $\mathcal{V}'_t$ one by one and detecting those that can be pruned by solving linear programs. Lark's algorithm works in a similar fashion except that its linear programs have fewer constraints. Since $\mathcal{V}'_t$ consists of $|\mathcal{A}||\hat{\mathcal{V}}_{t-1}|^{|\mathcal{O}|}$ state space functions, enumerating them one by one is very expensive. Other algorithms (Sondik, 1971; Cheng, 1988; Littman, 1994) avoid this difficulty by exploiting the structures of $\mathcal{V}'_t$.

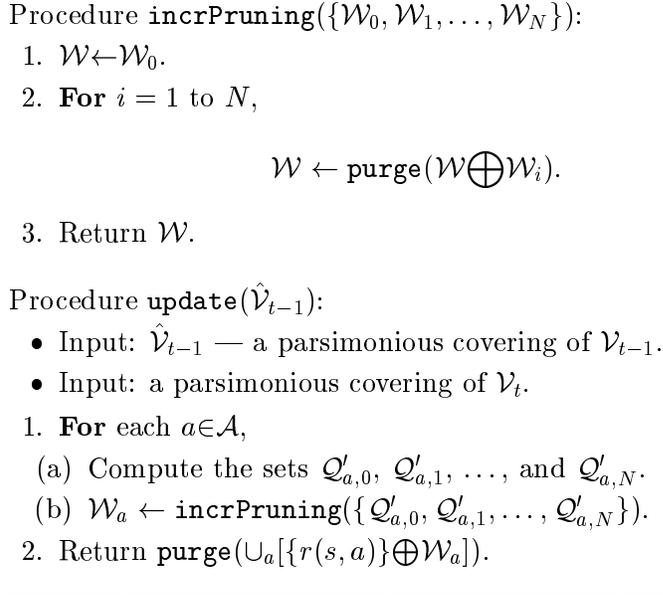

Figure 3: Incremental pruning and dynamic-programming updates.

## 5.4 A New Algorithm for Dynamic-Programming Updates

Let $\texttt{purge}(\mathcal{W})$ be a subroutine that takes a set $\mathcal{W}$ of state space functions and returns a parsimonious covering of $\mathcal{W}$. An implementation of $\texttt{purge}$ can be found in the appendix.

Let $\mathcal{W}_0, \ldots, \mathcal{W}_N$ be sets of state space functions. Consider the procedure $\texttt{incrPruning}$ shown in Figure 3. Let $n$ be a number between $0$ and $N$. Using Lemmas 1 and 2, one can easily show by induction that, at the end of the $n$th pass through the for-loop, the set $\mathcal{W}$ is a parsimonious covering of $\bigoplus_{i=0}^{n} \mathcal{W}_i$. Consequently, the procedures returns a parsimonious covering of $\bigoplus_{i=0}^{N} \mathcal{W}_i$. The procedure is named *incremental pruning* because pruning takes place after each cross sum.

Let $0, 1, \ldots, N$ be an enumeration of all the possible observations, i.e. possible instantiations of $o_+$. For any action $a$, suppose the sets $\mathcal{Q}'_{a,0}, \mathcal{Q}'_{a,1}, \ldots,$ and $\mathcal{Q}'_{a,N}$ have been computed. Applying $\texttt{incrPruning}$ to those sets, we get a parsimonious covering of $\bigoplus_{i=0}^{N} \mathcal{Q}'_{a,i}$. Denote it by $\mathcal{W}_a$. According to Lemma 2, $\cup_a[\{r(s,a)\}\bigoplus\mathcal{W}_a]$ covers $\mathcal{V}'_t$. Due to Lemma 1, this fact implies that a parsimonious covering of $\cup_a[\{r(s,a)\}\bigoplus\mathcal{W}_a]$ is also a parsimonious covering of $\mathcal{V}'_t$ and hence of $\mathcal{V}_t$. Thus, a parsimonious covering of $\mathcal{V}_t$ can be found from a parsimonious covering of $\mathcal{V}_{t-1}$ using the procedure $\texttt{update}$ shown in Figure 3.

We also use the term *incremental pruning* to refer to the above algorithm for dynamic-programming updates. It has been shown elsewhere (Cassandra *et al.*, 1977) that incremental pruning has the same asymptotic complexity as the witness algorithm and empirically it significantly outperforms the latter.

## 6. Region-Based Model Approximation

We have so far been concerned with exact algorithms. Experiments with incremental pruning, presently the most efficient exact algorithm, have revealed that it can solve only small POMDPs (Cassandra *et al.*, 1997). One needs to resort to approximation in order to solve large real-world problems.

Most previous approximation methods solve a POMDP directly; they approximate the *t*-step optimal value function of the POMDP. In the rest of this paper, we develop a new method that approximates a POMDP itself by another that has a more informative observation model and is hence easier to solve. The latter POMDP is solved and its solution is used to construct a solution to the original POMDP.

### 6.1 The Basic Idea

We make the following assumption about problem characteristics. in a POMDP $\mathcal{M}$, even though an agent does not know the true state of the world, it often has a good idea about the state. Justifications for this assumption were given in the introduction and empirical evidence is presented by Simmons & Koenig (1995).

Consider another POMDP $\mathcal{M}'$ that is the same as $\mathcal{M}$ except that in addition to the observation made by itself, the agent also receives a report from an oracle who knows the true state of the world. The oracle does not report the true state itself. Instead it selects, from a predetermined list of candidate regions, a region that contains the true state and reports that region.

More information is available to the agent in $\mathcal{M}'$ than in $\mathcal{M}$; additional information is provided by the oracle. Since in $\mathcal{M}$ the agent already has a good idea about the true state of the world, the oracle might not provide much additional information even when the candidate regions are small. Consequently, $\mathcal{M}'$ could be a good approximation of $\mathcal{M}$.

In $\mathcal{M}'$, the agent knows for sure that the true state of the world is in the region reported by the oracle. For this reason, we say that it is *region observable*. The region observable POMDP $\mathcal{M}'$ can be much easier to solve than $\mathcal{M}$ when the candidate regions are small. For example, if the oracle is allowed to report only singleton regions, then it actually reports the true state of the world and hence $\mathcal{M}'$ is an MDP. MDPs are much easier to solve than POMDPs. One would expect the region observable POMDP $\mathcal{M}'$ to be solvable when the candidate regions are small.

### 6.2 Spectrum of Approximations

If the region reported by the oracle is always the set of all possible states, then no additional information is provided, because the report that the true state of the world is one of the possible states has no information content. In this case, $\mathcal{M}'$ has the same solution as $\mathcal{M}$ and solving $\mathcal{M}'$ is equivalent to solving $\mathcal{M}$ directly. This is one extreme of the spectrum.

At the other extreme, if all the candidate regions are singletons, the oracle reports the true state of the world. Maximum amount of additional information is provided and $\mathcal{M}'$ is actually an MDP. The MDP might not be a good approximation of $\mathcal{M}$ but it is much easier to solve than $\mathcal{M}$.

Previous methods for solving a POMDP either solve it directly or to approximate it by using a MDP. By allowing the oracle to report regions that are neither singletons nor the set of all possible states, this paper opens up the possibility of exploring the spectrum between those two extremes. One way to explore the spectrum is to start with singleton candidate regions and increase their sizes gradually. Approximation quality and computational time both increase as one goes along. One stops when the approximation is accurate enough or the region observable POMDP becomes intractable. A method for determining approximation quality will be described later.

We now set out to make these ideas more concrete by starting with the concept of region systems.

## 6.3 Region Systems

A *region* is simply a subset of states of the world. A *region system* is a collection of regions such that no region is a subset of another region in the collection and the union of all regions equals the set of all possible states of the world. We use $R$ to denote a region and $\mathcal{R}$ to denote a region system.

Region systems are used to restrict the regions that the oracle can choose to report. The choice of a region system determines the computational complexity of the region observable POMDP $\mathcal{M}'$ and approximation quality. How to choose regions so as to make proper tradeoff between computational time and approximation quality is an open research issue. Here is a preliminary approach. The idea is to create a region for each state by including its "nearby" states. We say a state $s'$ is *ideally reachable in one step* from another state $s$ if after executing a certain action in state $s$, the probability of the world ending up in state $s'$ is the highest. A state $s_k$ is *ideally reachable in $k$ steps* from another state $s_0$ if there are state $s_1, \ldots, s_{k-1}$ such that $s_{i+1}$ is ideally reachable from $s_i$ in one step for all $0 \le i \le k-1$. Any state is ideally reachable from itself in 0 step.

For any non-negative integer $k$, the *radius-$k$ region centered at a state $s$* is the set of states that are ideally reachable from $s$ in $k$ or less steps. A *radius-$k$ region system* is the one obtained by creating a radius-$k$ region for each state and then removing, one after another, regions that are subsets of others.

When $k$ is 0, the radius-$k$ region system consists of singleton regions. On the other hand, if each state is reachable from any other state in $k$ or less steps, there is only one region in the radius-$k$ region system — the set of all possible states.

## 6.4 Region Observable POMDPs

Given a region system $\mathcal{R}$ and a POMDP $\mathcal{M}$, we construct a region observable POMDP $\mathcal{M}'$ by assuming that at each time point the agent not only obtains an observation by itself but also receives a report from an oracle who knows the true state of the world. The oracle does not report the true state itself. Instead it chooses from $\mathcal{R}$ one region that contains the true state and reports that region.

The amount of additional information provided by the oracle depends not only on the region system used but also on the way the oracle chooses regions. For example, if the oracle always reports the region centered at the true state, then it implicitly reports the true state itself.

In order to provide as little additional information as possible, the oracle should consider what the agent already knows. However, it cannot take the entire history of past actions and observations into account because if it did, $\mathcal{M}'$ would not be a POMDP. The current observation would depend on the entire history.

For any non-negative state space function $f(s)$ and any region $R$, we call the quantity $supp(f, R) = \sum_{s \in R} f(s) / \sum_{s \in \mathcal{S}} f(s)$ the *degree of support* of $f$ by $R$. Note that when $f$ is a probability distribution, the denominator is 1. If $R$ supports $f$ to degree 1, we say that $R$ *fully supports* $f$.

We suggest the following region-selection rule for the oracle. Let $s_-$ be the previous true state of the world, $a_-$ be the previous action, and $o$ be the current observation. The oracle should choose, among all the regions in $\mathcal{R}$ that contain the true state of the world, one that supports the function $P(s, o|s_-, a_-)$ of $s$ to the maximum degree. Where there is more than one such regions, choose the one that comes first in a predetermined ordering among the regions.

Here are some arguments in support of the rule. If the previous world state $s_-$ were known to the agent, then its current belief state $b(s)$, a function of s, would be proportional to $P(s, o|s_-, a_-)$. In this case, the rule minimizes additional information in the sense that the region reported supports the current belief state to the maximum degree. If the previous world state is known to be around $s_-$, the same is roughly true. Also if the current observation is informative enough, being a landmark for instance, to ensure that the world state is in a certain region, then the region chosen using the rule fully supports the current belief state. In such a case, no additional information is provided. Despite those arguments, we do not claim that the rule described above is optimal. Finding a rule that minimizes additional information is still an open problem.

The probability $P(R|s, o, s_-, a_-)$ of a region $R$ being chosen under the above scheme is given by

$$P(R|s, o, s_-, a_-) = \begin{cases} 1 & \text{if } R \text{ is the first region s.t. } s \in R \text{ and for any other region } R' \\ & \sum_{s' \in R} P(s', o|s_-, a_-) \geq \sum_{s' \in R'} P(s', o|s_-, a_-) \\ 0 & \text{otherwise.} \end{cases}$$

The region observable POMDP $\mathcal{M}'$ differs from the original POMDP $\mathcal{M}$ only in terms of observation; in addition to the observation $o$ made by itself, the agent also receives a report $R$ from the oracle. We shall denote an observation in $\mathcal{M}'$ by $z$ and write $z = (o, R)$. The observation model of $\mathcal{M}'$ is given by

$$P(z|s, a_-, s_-) = P(o, R|s, a_-, s_-) = P(o|s, a_-)P(R|s, o, s_-, a_-).$$

The joint conditional probability $P(s_+, z_+|s, a)$ of the next state $s_+$ of the world and the next observation $z_+$ given the current state $s$ and the current action $a$ is

$$P(s_+, z_+|s, a) = P(s_+|s, a)P(z_+|s_+, a, s).$$

## 7. Solving Region Observable POMDPs

In principle, the region observable POMDP $\mathcal{M}'$ can be solved in the same way as general POMDPs using the procedure `solvePOMDP`. It is not advisable to do so, however, since `solvePOMDP` does not automatically exploit region observability. This section and the next two sections develop an algorithm for solving $\mathcal{M}'$ that takes advantage of region observability.

### 7.1 Restricted Value Iteration

For any region $R$, let $\mathcal{B}_R$ be the set of belief states that are fully supported by $R$. Let $\mathcal{R}$ be the region system underlying the region observable POMDP $\mathcal{M}'$. Define $\mathcal{B}_{\mathcal{R}} = \cup_{R \in \mathcal{R}} \mathcal{B}_R$.

It is easy to see that, in $\mathcal{M}'$, no matter what the current belief state $b$ is, the next belief state $b_+$ must be in $\mathcal{B}_{\mathcal{R}}$. We assume that the initial belief state is in $\mathcal{B}_{\mathcal{R}}$. Then all possible belief states the agent might encounter are in $\mathcal{B}_{\mathcal{R}}$. This implies that policies for $\mathcal{M}'$ need only be defined over $\mathcal{B}_{\mathcal{R}}$ and value iteration for $\mathcal{M}'$ can be restricted to the subset $\mathcal{B}_{\mathcal{R}}$ of $\mathcal{B}$.

We restrict value iteration for $\mathcal{M}'$ to $\mathcal{B}_{\mathcal{R}}$ for the sake of efficiency. Doing so implies that the $t$-step optimal value function of $\mathcal{M}'$, denoted by $U_t^*$, is defined only over $\mathcal{B}_{\mathcal{R}}$ and the Bellman residual is now $max_{b \in \mathcal{B}_{\mathcal{R}}} |U_t^*(b) - U_{t-1}^*(b)|$. To avoid confusion, we call it the *restricted Bellman residual* and call $U_t^*$ the *restricted $t$-step optimal value function*.

Since $\mathcal{B}_{\mathcal{R}}$ is continuous, restricted value iteration cannot be carried out implicitly. The next subsection shows how it can be carried implicitly.

### 7.2 Implicit Restricted Value Iteration

Let $\mathcal{W}$ and $\mathcal{X}$ be two sets of state space functions and let $R$ be a region. We say that $\mathcal{W}$ *covers* $\mathcal{X}$ *in region* $R$ if, for any $b \in \mathcal{B}_R$,

$$\mathcal{W}(b) = \mathcal{X}(b).$$

We say that $\mathcal{W}$ *parsimoniously covers* $\mathcal{X}$ *in region* $R$ if $\mathcal{W}$ covers $\mathcal{X}$ in region $R$ and none of its proper subsets do. When $\mathcal{W}$ covers or parsimoniously covers $\mathcal{X}$ in a region, we refer to $\mathcal{W}$ as a *regional covering* or a *parsimonious regional covering* of $\mathcal{X}$.

Let $\mathcal{U}_t$ be the set of all $t$-step state value functions of $\mathcal{M}'$. According to Theorem 2,

$$U_t^*(b) = \mathcal{U}_t(b)$$

for any belief state $b \in \mathcal{B}_{\mathcal{R}}$.

For each region $R$, suppose $\hat{\mathcal{U}}_{t,R}$ is a set of state space functions that parsimoniously covers $\mathcal{U}_t$ in region $R$. Then the collection $\{\hat{\mathcal{U}}_{t,R} | R \in \mathcal{R}\}$ is a *representation* of $U_t^*$ in the sense that for any $b \in \mathcal{B}_{\mathcal{R}}$,

$$U_t^*(b) = \hat{\mathcal{U}}_{t,R_b}(b), \tag{12}$$

where $R_b$ is such that $b \in \mathcal{B}_R$, i.e. such that $R_b$ fully supports $b$.

As will be shown in the next section, parsimonious regional coverings of $\mathcal{U}_t$ can be obtained from parsimonious regional coverings of $\mathcal{U}_{t-1}$. Let `ROPOMDPupdate`$(R, \hat{\mathcal{U}}_{t-1,R_+} | R_+ \in \mathcal{R}\})$ be a procedure that takes a region $R$ and parsimonious regional covering $\{\hat{\mathcal{U}}_{t-1,R_+} | R_+ \in \mathcal{R}\}$

––––––––––––––––––––––––––––––––––––––––––––––––––––

Procedure `solveROPOMDP`($\mathcal{M}'$, $\epsilon$)

- Input: $\mathcal{M}'$ — A region observable POMDP,
   $\epsilon$ — A positive number.

- Output: A list of sets of state space functions.

1. $t \leftarrow 0$.
2. **For** $R \in \mathcal{R}$, $\hat{\mathcal{U}}_{0,R} \leftarrow \{0\}$.
3. **Do**
   - $t \leftarrow t+1$.
   - **For** each $R \in \mathcal{R}$,

      $\hat{\mathcal{U}}_{t,R} \leftarrow$ `ROPOMDPupdate`($R$, $\{\hat{\mathcal{U}}_{t-1,R_+}|R_+ \in \mathcal{R}\}$).

   **while** `ROPOMDPstop`($\{\hat{\mathcal{U}}_{t,R}|R \in \mathcal{R}\}$, $\{\hat{\mathcal{U}}_{t-1,R_+}|R_+ \in \mathcal{R}\}$, $\epsilon$) = `no`.
4. Return $\{\hat{\mathcal{U}}_{t,R}|R \in \mathcal{R}\}$.

––––––––––––––––––––––––––––––––––––––––––––––––––––

Figure 4: Implicit restricted value iteration for region-observable POMDPs.

of $\mathcal{U}_{t-1}$ and returns a set of state space functions that parsimoniously covers $\mathcal{U}_t$ in region $R$ [2]. Let `ROPOMDPstop` be a procedure that determines, from parsimonious regional coverings of $\mathcal{U}_{t-1}$ and $\mathcal{U}_t$, whether the restricted Bellman residual has fallen below a predetermined threshold.

The procedure `solveROPOMDP` shown in Figure 4 carries out restricted value iteration implicitly: instead inductively computing the restricted $t$-step optimal value function $U_t^*$ itself, it computes parsimonious regional coverings of $\mathcal{U}_t$. In other words, it computes sets of state space functions that represent $U_t^*$ in the sense of (12).

Let $\pi'$ be the greedy policy for $\mathcal{M}'$ based on $U_t^*$. For any $b \in \mathcal{B}_\mathcal{R}$, $\pi'(b)$ is defined by Equation (7) with $o_+$ replaced by $z_+ = (o_+, R_+)$ and $V_t^*$ replaced by $U_t^*$. Since the list $\{\hat{\mathcal{U}}_{t,R}|R \in \mathcal{R}\}$ of sets of state space functions returned by `solveROPOMDP` represents $U_t^*$ in the sense of (12), we have that for any $b \in \mathcal{B}_\mathcal{R}$

$$\pi'(b) = arg\ max_a [r(b,a) + \gamma \sum_{o_+, R_+} P((o_+, R_+)|b, a)\hat{\mathcal{U}}_{t,R_+}(b_+)]. \qquad (13)$$

The next two sections show how to implement the procedures `ROPOMDPupdate` and `ROPOMDPstop`.

––––––––––––––––––––––––––––––

2. The string "ROPOMDP" in `ROPOMDPupdate` stands for region-observable POMDP.

## 8. Dynamic-Programming Updates for Region Observable POMDPs

This section shows how the incremental pruning algorithm developed in Section 5 can be adapted to compute parsimonious regional coverings of $\mathcal{U}_t$ from parsimonious regional coverings of $\mathcal{U}_{t-1}$.

### 8.1 Properties of Regional Coverings

**Lemma 3** *Let $R$ be a region and let $\mathcal{W}$, $\mathcal{X}$, and $\mathcal{Y}$ be three sets of state space functions. If $\mathcal{W}$ parsimoniously covers $\mathcal{X}$ in region $R$ and $\mathcal{X}$ covers $\mathcal{Y}$ in region $R$, then $\mathcal{W}$ parsimoniously covers $\mathcal{Y}$ in region $R$.* $\square$

**Lemma 4** *Let $R$ be a region and let $\mathcal{W}$, $\mathcal{W}'$, $\mathcal{X}$, and $\mathcal{X}'$ be four sets of state space functions. If $\mathcal{W}'$ and $\mathcal{X}'$ respectively cover $\mathcal{W}$ and $\mathcal{X}$ in region $R$, then*

1. *$\mathcal{W}'\bigoplus\mathcal{X}'$ covers $\mathcal{W}\bigoplus\mathcal{X}$ in region $R$.*

2. *$\mathcal{W}'\cup\mathcal{X}'$ covers $\mathcal{W}\cup\mathcal{X}$ in region $R$.* $\square$

### 8.2 Regional Coverings of $\mathcal{U}_t$ from Parsimonious Regional Coverings of $\mathcal{U}_{t-1}$

From parsimonious regional coverings $\hat{\mathcal{U}}_{t-1,R_+}$ $(R_+\in\mathcal{R})$ of $\mathcal{U}_{t-1}$, this subsection constructs, for each region $R\in\mathcal{R}$, a set $\mathcal{U}_{t,R}$ of state space functions and shows that it covers $\mathcal{U}_t$ in region $R$.

For any action $a$ and any observation $z_+=(o_+, R_+)$ of $\mathcal{M}'$, let $\mathcal{Q}_{a,z_+,R}$ be the set of all state space functions $\beta$ that are of the following form:

$$\beta(s) = \begin{cases} \gamma \sum_{s_+} \alpha(s_+)P(s_+, z_+|s, a) & \text{if } s\in R \\ 0 & \text{otherwise.} \end{cases} \tag{14}$$

where $\alpha \in \hat{\mathcal{U}}_{t-1,R_+}$. Define

$$\mathcal{U}_{t,R} \;=\; \cup_a[\{r(s,a)\}\bigoplus(\bigoplus_{z_+}\mathcal{Q}_{a,z_+,R})].$$

**Proposition 3** *The set $\mathcal{U}_{t,R}$ covers $\mathcal{U}_t$ in region $R$.*

Formal proof of this proposition can be found in Appendix A. Informally, the fact that $\hat{\mathcal{U}}_{t-1,R_+}$ covers $\mathcal{U}_{t-1}$ in region $R_+$ implies that $\mathcal{Q}_{a,z_+,R}$ covers $\mathcal{Q}_{a,z_+}$ in region $R$, where $\mathcal{Q}_{a,z_+}$ is given by (11) with $o_+$ and $\mathcal{V}_{t-1}$ replaced by $z_+$ and $\mathcal{U}_{t-1}$. This fact in turn implies that $\mathcal{U}_{t,R}$ covers $\mathcal{U}_t$ in region $R$ because of Proposition 1 and Lemma 4.

### 8.3 Possible Observations at the Next Time Point

In the definition of $\mathcal{U}_{t,R}$, the cross sum is taken over all possible observations. This subsection shows that some of the possible observations can be skipped.

For any action $a$ and any region $R$, define

$$Z_{a,R} = \{z_+|\sum_{s_+} P(s_+, z_+|s, a) > 0 \text{ for some } s\in R \}.$$

_______________________________________________

Procedure `ROPOMDPupdate`$(R, \hat{\mathcal{U}}_{t-1,R_+} | R_+ \in \mathcal{R}\})$:

- Inputs: $R$ — A region, and for any region $R_+$,
    $\hat{\mathcal{U}}_{t-1,R_+}$ parsimoniously covers $\mathcal{U}_{t-1}$ in region $R_+$.

- Output: A set of state space functions that parsimoniously covers $\mathcal{U}_t$ in region $R$.

1. **For** each action $a$,

    (a) Compute the set $\mathcal{Z}_{a,R}$ and enumerate its members as $0, 1, \ldots, M$.

    (b) **For** $i=0$ to $M$, compute the set $\mathcal{Q}_{a,i,R}$.

    (c) $\mathcal{W}_a \leftarrow$ `restrictedIncrPruning`$(\{\mathcal{Q}_{a,0,R}, \mathcal{Q}_{a,1,R}, \ldots, \mathcal{Q}_{a,M,R}\}, R)$.

2. Return `purge`$(\cup_a[\{r(s,a)\} \bigoplus \mathcal{W}_a], R)$.

Subroutine `restrictedIncrPruning`$(\{\mathcal{W}_0, \mathcal{W}_1, \ldots, \mathcal{W}_M\}, R)$:

1. Let $\mathcal{W} \leftarrow \mathcal{W}_0$.

2. **For** $i=1$ to $M$,
$$\mathcal{W} \leftarrow \texttt{purge}(\mathcal{W} \bigoplus \mathcal{W}_i, R).$$

3. Return $\mathcal{W}$.

_______________________________________________

Figure 5: Dynamic-programming updates for region observable POMDPs.

It is the set of observations that the agent can possibly receive at the next time point given that the current state of the world lies in region $R$ and the current action is $a$. There are many observations outside this set. As a matter of fact, an observation $z_+ = (o_+, R_+)$ is not in the set if it is not possible to reach region $R_+$ from region $R$ in one step.

For any $z_+ = (o_+, R_+)$, if $z_+ \notin Z_{a,R}$, then $\sum_{s_+} P(s_+, z_+ | s, a) = 0$ for all $s \in R$. In such a case, $\mathcal{Q}_{a,z_+,R} = \{0\}$ according to (14). Since, $\{0\} \bigoplus \mathcal{W} = \mathcal{W}$ for any set $\mathcal{W}$ of state space functions, we have
$$\mathcal{U}_{t,R} = \cup_a[\{r(s,a)\} \bigoplus (\bigoplus_{z_+ \in Z_{a,R}} \mathcal{Q}_{a,z_+,R})].$$

## 8.4 Parsimonious Regional Covering of $\mathcal{U}_t$

Proposition 3 and Lemma 3 imply that, for any region $R$, a set of state space functions parsimoniously covers $\mathcal{U}_t$ in region $R$ if and only if it parsimoniously covering $\mathcal{U}_{t,R}$ in region $R$. According to Lemmas 3 and 4, a set of state space functions that parsimoniously covers $\mathcal{U}_{t,R}$ in region $R$ can be found using the procedure `ROPOMDPupdate` shown in Figure 5 (c.f. Section 5.4). In the procedure, the subroutine `purge`$(\mathcal{W}, R)$ takes a set $\mathcal{W}$ of state space

———————————————————————————————————

Procedure `ROPOMDPstop`($\{\hat{\mathcal{U}}_{t,R}|R \in \mathcal{R}\}, \{\hat{\mathcal{U}}_{t-1,R}|R \in \mathcal{R}\}, \epsilon$)

- Inputs: $\epsilon$ — A positive number, and for any region $R$
  $\hat{\mathcal{U}}_{t,R}$ covers $\mathcal{U}_t$ in region $R$, and
  $\hat{\mathcal{U}}_{t-1,R}$ covers $\mathcal{U}_{t-1}$ in region $R$.
- Outputs: `yes` — If the restricted Bellman residual $\leq \epsilon$,
  `no` — Otherwise.

1. **For** each region $R$,
   (a) flag $\leftarrow$ `yes`.
   (b) **For** each $\alpha \in \hat{\mathcal{U}}_{t-1,R}$,

   flag $\leftarrow$ `no` if `dominate`$(\alpha, \hat{\mathcal{U}}_{t,R}, R, \epsilon) \neq$ `nil`.

   (c) **For** each $\alpha \in \hat{\mathcal{U}}_{t,R}$,

   flag $\leftarrow$ `no` if `dominate`$(\alpha, \hat{\mathcal{U}}_{t-1,R}, R, \epsilon) \neq$ `nil`.

   (d) Return `no` if flag = `no`.
2. Return `yes`.

———————————————————————————————————

Figure 6: Procedure for determining whether the restricted Bellman residual has fallen below a threshold.

functions and region $R$, and returns a set of state space functions that parsimoniously covers $\mathcal{W}$ in region $R$. An implementation of this subroutine can be found in Appendix B.

## 9. The Stopping Condition

This section shows how to determine whether the restricted Bellman residual has fallen below a predetermined threshold $\epsilon$ from regional coverings of $\mathcal{U}_t$ and $\mathcal{U}_{t-1}$. For any region $R$, let $\hat{\mathcal{U}}_{t,R}$ and $\hat{\mathcal{U}}_{t-1,R}$ be two sets of state space functions that respectively cover $\mathcal{U}_t$ and $\mathcal{U}_{t-1}$ in region $R$. By the definition of regional coverings, we have

**Lemma 5** *The restricted Bellman residual is no larger than $\epsilon$ if and only if for any region $R$ and any belief state $b \in \mathcal{B}_R$,*

1. *For any $\alpha \in \hat{\mathcal{U}}_{t,R}$,*

$$\alpha(b) \leq \hat{\mathcal{U}}_{t-1,R}(b) + \epsilon, \ \ and$$

2. *For any $\alpha \in \hat{\mathcal{U}}_{t-1,R}$,*

$$\alpha(b) \leq \hat{\mathcal{U}}_{t,R}(b) + \epsilon. \ \Box$$

Let `dominate`$(\alpha, \mathcal{W}, R, \epsilon)$ be a procedure that returns a belief state $b$ in $\mathcal{B}_R$ such that $\alpha(b) > \mathcal{W}(b) + \epsilon$. If such a belief state does not exist, it returns `nil`. An implementation of this procedure can be found in Appendix B. The procedure `ROPOMDPupdate` shown in Figure 6 returns `yes` if the restricted Bellman residual has fallen below $\epsilon$ and `no` otherwise.

A couple of notes are in order. First, when the reward function $r(s, a)$ is non-negative, $U_t^*$ increases with $t$. In this case, the restricted Bellman residual becomes $max_{b \in \mathcal{B}_R}(U_t^*(b) - U_{t-1}^*(b))$. Consequently, step (c) can be skipped. Second, when $r(s, a)$ takes negative values for some $s$ and some $a$, a constant can be added to it so that it becomes non-negative. Adding a constant to $r(s, a)$ does not affect the optimal policy. However, it makes it easier to determine whether the restricted Bellman residual has fallen below a threshold.

## 10. Decision Making for the Original POMDP

Suppose we have solved the region observable POMDP $\mathcal{M}'$. The next step is to construct a policy $\pi$ for the original POMDP $\mathcal{M}$ based on the solution for $\mathcal{M}'$.

Even though it is our assumption that in the original POMDP $\mathcal{M}$ the agent has a good idea about the state of the world at all times, there is no guarantee that its belief state are always in $\mathcal{B}_{\mathcal{R}}$. There is no oracle in $\mathcal{M}$. A policy should prescribe actions for belief states in $\mathcal{B}_{\mathcal{R}}$ as well as for belief states outside $\mathcal{B}_{\mathcal{R}}$. One issue here is that the policy $\pi'$ for $\mathcal{M}'$ is defined only for belief states in $\mathcal{B}_{\mathcal{R}}$. Fortunately, $\pi'$ can be naturally extended to the entire belief space by ignoring the constraint $b \in \mathcal{B}_{\mathcal{R}}$ in Equation (13). We hence define a policy $\pi$ for $\mathcal{M}$ as follows: for any $b \in \mathcal{B}$,

$$\pi(b) = arg\ max_a[r(b, a) + \gamma \sum_{o_+, R_+} P((o_+, R_+)|b, a)\hat{\mathcal{U}}_{t, R_+}(b_+)]. \tag{15}$$

Let $k$ be the radius of the region system underlying $\mathcal{M}'$. The policy $\pi$ given above will be referred to as the *radius-k approximate policy* for $\mathcal{M}$. The entire process of obtaining this policy, including the construction and solving of the region observable POMDP $\mathcal{M}'$, will be referred to as *region-based approximation*.

It is worthwhile to compare this equation with Equation (7). In Equation (7), there are two terms on the right hand side. The first term is the immediate reward for taking action $a$ and the second term is the discounted future reward the agent can expect to receive if it behaves optimally. Their sum is the total expected reward for taking action $a$. The action with the highest total reward is chosen.

The second term is difficult to obtain. In essence, Equation (15) approximates the second term using the optimal expected future reward the agent can receive with the help of the oracle, which is easier to compute.

It should be emphasized that the presence of the oracle is assumed only in the process of computing the radius-k approximate policy. The oracle is not present when executing the policy.

## 11. Quality of Approximation and Simulation

In general, the quality of an approximate policy $\pi$ is measured by the distance between its value function $V^\pi(b)$ and the optimal value function $V^*(b)$. This measurement does not consider what an agent might know about the initial state of the world. As such, it is not appropriate for a policy obtained through region-based approximation. One cannot expect such a policy to be of good quality if an agent is very uncertain about the initial state of the world because it is obtained under the assumption that an agent has a good idea about the state of the world at all times.

This section describes a scheme for determining the quality of an approximate policy in cases where an agent knows the initial state of the world with certainty. The scheme can be generalized to cases where there is a small amount of uncertainty about the initial state; for example, cases where the initial state is known to be in some small region.

An agent might need to reach a goal from different initial states at different times. Let $P(s)$ be the frequency it will start from state $s$[3]. The quality of an approximate policy $\pi$ can be measured by $\sum_s |V^*(s) - V^\pi(s)|P(s)$, where $V^*(s)$ and $V^\pi(s)$ denote the rewards the agent can expect to receive starting from state $s$ if it behaves optimally or according to $\pi$ respectively.

By definition $V^*(s) \geq V^\pi(s)$ for all $s$. Let $U^*$ be the optimal value function of the region observable POMDP $\mathcal{M}'$. Since more information is available to the agent in $\mathcal{M}'$, $U^*(s) \geq V^*(s)$ for all $s$. Therefore, $\sum_s [U^*(s) - V^\pi(s)]P(s)$ is an upper bound on $\sum_s [V^*(s) - V^\pi(s)]P(s)$.

Let $\pi'$ be the policy for $\mathcal{M}'$ given by (13). When the restricted Bellman residual is small, $\pi'$ is close to optimal for $\mathcal{M}'$ and the value function $V^{\pi'}$ of $\pi'$ is close to $U^*$. Consequently, $\sum_s [V^{\pi'}(s) - V^\pi(s)]P(s)$ is an upper bound on $\sum_s [V^*(s) - V^\pi(s)]P(s)$ when the restricted Bellman residual is small enough.

One way to estimate the quantity $\sum_s [V^{\pi'}(s) - V^\pi(s)]P(s)$ is to conduct a large number of simulation trials. In each trial, an initial state is randomly generated according to $P(s)$. The agent is informed of the initial state. Simulation takes place in both $\mathcal{M}$ and $\mathcal{M}'$. In $\mathcal{M}$, the agent chooses, at each step, an action using $\pi$ based on its current belief state. The action is passed to a simulator which randomly generates the next state of the world and the next observation according to the transition and observation probabilities. The observation (but not the state) is passed to the agent, who updates its belief state and chooses the next action. And so on and so forth. The trial terminates when the agent chooses the action `declare-goal` or a maximum number of steps is reached. Simulation in $\mathcal{M}'$ takes place in a similar manner except that the observations and the observation probabilities are different and actions are chosen using $\pi'$.

If the goal is correctly declared at the end of a trial, the agent receives a reward in the amount $\gamma^n$, where $n$ is the number of steps. Otherwise, the agent receive no reward. The quantity $\sum_s [V^{\pi'}(s) - V^\pi(s)]P(s)$ can be estimated using the difference between the average reward received in the trials for $\mathcal{M}'$ and the average reward received in the trials for $\mathcal{M}$.

## 12. Tradeoff Between Quality and Complexity

Intuitively, the larger the radius of the region system, the less the amount of additional information the oracle provides. Hence the closer $\mathcal{M}'$ is to $\mathcal{M}$ and the narrower the gap between $\sum_s V^{\pi'}(s)P(s)$ and $\sum_s V^\pi(s)P(s)$. Although we have not theoretically proved this, empirical results (see the next section) do suggest that $\sum_s V^\pi(s)P(s)$ increases with the radius of the region system while $\sum_s V^{\pi'}(s)P(s)$ decreases with it. In the extreme case when there is one region in the region system that contains all the possible states of the world, $\mathcal{M}$ and $\mathcal{M}'$ are identical and hence so are $\sum_s V^{\pi'}(s)P(s)$ and $\sum_s V^\pi(s)P(s)$.

---

3. This is not to be confused with the initial belief state $P_0$, which represents the agent's knowledge about the ninitial state at a particular trial.

These discussions lead to the following scheme for making the tradeoff between complexity and quality: Start with the radius-0 region system and increase the radius gradually until the quantity $\sum_s [V^{\pi'}(s) - V_\pi(s)]P(s)$ becomes sufficiently small or the region observable POMDP $\mathcal{M}'$ becomes untractable.

## 13. Simulation Experiments

Simulation experiments have been carried out to show that (1) approximation quality increases with radius of region system and (2) where there is not much uncertainty, a POMDP can be accurately approximated by a region-observable POMDP that can be solved exactly. This section reports on the experiments.

### 13.1 Synthetic Office Environments

Our experiments were conducted using two synthetic office environments borrowed from Cassandra *et al.* (1996) with some minor modifications. Layouts of the environments are shown in Figure 7, where squares represent locations. Each location is represented as four states in the POMDP model, one for each orientation. The dark locations are rooms connected to corridors by doorways.

In each environment, a robot needs to reach the goal location with the correct orientation. At each step, the robot can execute one of the following actions: `move-forward`, `turn-left`, `turn-right`, and `declare-goal`. The two sets of action models given in Figure 7 were used in our experiments. For the action `move-forward`, the term F-F (0.01) means that with probability 0.01 the robot actually moves two steps forward. The other terms are to be interpreted similarly. If an outcome cannot occur in a certain state of the world, then the robot is left in the last state before the impossible outcome.

In each state, the robot is able to perceive in each of three nominal directions (front, left, and right) whether there is a `doorway`, `wall`, `open`, or it is `undetermined`. The two sets of observation models shown in Figure 7 were used in our experiments.

### 13.2 Complexity of Solving the POMDPs

One of the POMDPs has 280 possible states while the other has 200. They both have 64 possible observations and 4 possible actions. Since the largest POMDPs that researchers have been able to solve exactly so far have less than 20 states and 15 observations, it is safe to say no existing exact algorithms can solve those two POMDPs.

We were able to solve the radius-0 and radius-1 approximations (region observable POMDPs) of the two POMDPs on a SUN SPARC20 computer. The threshold for the Bellman residual was set at 0.001 and the discount factor at 0.99. The amounts of time it took in CPU seconds are collected in the following table.

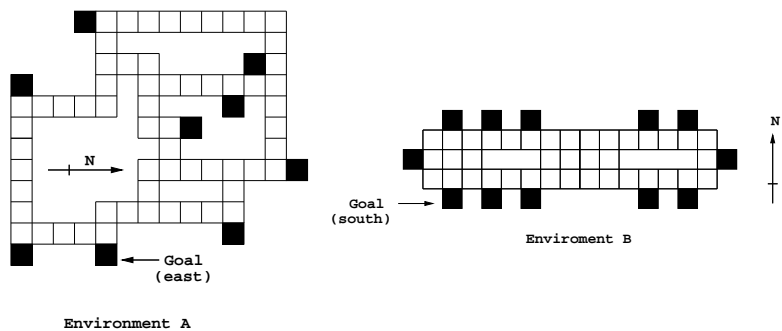

**Transition Probabilities**

| Action | Standard outcomes | Noisy outcomes |
|---|---|---|
| `move-forward` | N (0.11), F (0.88), F-F (0.01) | N (0.2), F (0.7), F-F (0.1) |
| `turn-left` | N (0.05), L (0.9), L-L (0.05) | N (0.15), L (0.7), L-L (0.15) |
| `turn-right` | N (0.05), R (0.9), R-R (0.05) | N (0.15), R (0.7), R-R (0.15) |
| `declare-goal` | N (1.0) | N (1.0) |

**Observation Probabilities**

| Actual case | Standard observations | Noisy observations |
|---|---|---|
| `wall` | wall (0.90), open (0.04), doorway (0.04), undetermined (0.02) | wall (0.70), open (0.19), doorway (0.09), undetermined (0.02) |
| `open` | wall (0.02), open (0.90), doorway (0.06), undetermined (0.02) | wall (0.19), open (0.70), doorway (0.09), undetermined (0.02) |
| `doorway` | wall (0.15), open (0.15), doorway (0.69), undetermined (0.01) | wall (0.15), open (0.15), doorway (0.69), undetermined (0.01) |

Figure 7: Synthetic Office Environments.

| Environment | Standard models | | Noisy models | |
|---|---|---|---|---|
| | Radius-0 | Radius-1 | Radius-0 | Radius-1 |
| A | 1.26 | 3373 | 1.35 | 5984 |
| B | 0.61 | 2437 | 0.72 | 3952 |

We see that the radius-1 approximations took much longer time to solve than the radius-0 approximations. Also notice that the region observable POMDPs with noisy action and observation models took more time to solve that those with the standard models. This suggests that the more nondeterministic the actions and the less informative the observations, the more difficult it is to solve a POMDP.

We were unable to solve the radius-2 approximations. Other approximation techniques need to be incorporated in order to solve the approximations based on region systems with radius larger than or equal to 2.

## 13.3 Approximation Quality for Standard Models

To determine the quality of the radius-0 and radius-1 approximate policies for the POMDPs with standard action and observation models, 1000 simulation trials were conducted using the scheme described in Section 11. It was assumed that the agent is equally likely to start from any state. Average rewards obtained in the original POMDPs $\mathcal{M}$ (i.e. without the help of the oracle) and in the corresponding region-observable POMDPs $\mathcal{M}'$ (i.e. with the help of the oracle) are shown in the following table.

| | Environment A | | Environment B | |
| --- | --- | --- | --- | --- |
| | radius-0 | radius-1 | radius-0 | radius-1 |
| Average reward in $\mathcal{M}$ | 0.806535 | 0.815695 | 0.866118 | 0.868533 |
| Average reward in $\mathcal{M}'$ | 0.827788 | 0.818534 | 0.883271 | 0.876356 |
| Difference | 0.021253 | 0.002839 | 0.017153 | 0.007823 |

We see that, when the radius-0 policies were used, the differences between the rewards obtained in $\mathcal{M}$ and those obtained in $\mathcal{M}'$ are very small in both environments. This indicates that the radius-0 region observable POMDPs (i.e. MDPs) are accurate approximations of the original POMDPs. The radius-0 approximate policies are close to optimal for the original POMDPs. When the radius-1 policies were used, the differences are even smaller; the rewards obtained in $\mathcal{M}$ and those obtained in $\mathcal{M}'$ are essentially the same.

Consider the rewards obtained in the original POMDPs. We see that they are larger when radius-1 policies were used than when radius-0 policies were used. This supports our claim that approximation quality increases with radius of region system.

There is a another fact worth mentioning. The differences between rewards obtained in $\mathcal{M}$ and those obtained in $\mathcal{M}'$ are larger in Environment B than in Environment A. This is because Environment B is more symmetric and consequently observations are less effective in disambiguating uncertainty in the agent's belief about the state of the world.

## 13.4 Approximation Quality for Noisy Models

One thousand trials were also conducted for the POMDPs with noisy action and observation models. Results are shown in the following table.

| | Environment A | | Environment B | |
| --- | --- | --- | --- | --- |
| | radius-0 | radius-1 | radius-0 | radius-1 |
| Average reward in $\mathcal{M}$ | 0.596670 | 0.634934 | 0.445653 | 0.565099 |
| Average reward in $\mathcal{M}'$ | 0.812898 | 0.722441 | 0.871903 | 0.818365 |
| Difference | 0.214228 | 0.087507 | 0.426250 | 0.253266 |

We see that the differences between rewards obtained in $\mathcal{M}$ and rewards obtained in $\mathcal{M}'$ are significantly smaller when the radius-1 policies were used than when the radius-0 policies were used. This is the case especially in Environment A. Also the rewards obtained in $\mathcal{M}$ are larger when the radius-1 policies were used than when the radius-0 policies were used. Those again support our claim that approximation quality increases with radius of region system.

As far as absolute approximation quality is concerned, the radius-0 POMDPs (i.e. MDPs) are obviously very poor approximations of the original POMDPs; when the radius-0 policies were used, the rewards obtained in $\mathcal{M}$ are significantly smaller than the rewards obtained in $\mathcal{M}'$. For Environment A, the radius-1 approximation is fairly accurate. However, the radius-1 approximation remains poor for Environment B. The radius of region system needs to be increased.

Tracing through the trials step by step, we observed some interesting facts. In Environment B, the agent, under the guidance of the radius-1 approximate policy, was able to quickly get to the neighborhood of the goal even when starting from far way. The fact that the Environment around the goal is highly symmetric was the cause of the poor performance. Often the agent was not able to determine whether it was at the goal location (room), or in the opposite room, or in the left most room, or in the room to the right of the goal location. The performance would be close to optimal if the goal location had some distinct features.

In Environment A, the agent, again under the guidance of the radius-1 approximate policy, was able to reach and declare the goal successfully once it got to the neighborhood. However, it often took many unnecessarily steps before reaching the neighborhood due to uncertainty in the effects of the turning actions. For example, when the agent reached the lower left corner from above, it was facing downward. The agent executed the action `turn_left`. Fifteen percent of the time, it ended up facing upward instead of to the right. The agent then decided to `move-forward`, thinking that it was approaching the goal. But it was actually moving upward and did not realize this until a few steps later. The agent would perform much better if there were informative landmarks around the corners.

## 14. Conclusions

We propose to approximate a POMDP by using a region observable POMDP. The region observable POMDP has more informative observations and hence is easier to solve. A method for determining approximation quality is described, which allows one to make the tradeoff between approximation quality and computational time by starting with a coarse approximation and refining it gradually. Simulation experiments have shown that when there is not much uncertainty in the effects of actions and observations are informative, a POMDP can be accurately approximated by a region observable POMDP that can be solved exactly. However, this becomes infeasible as the degree of uncertainty increases. Other approximate methods need to be incorporated in order to solve region observable POMDPs whose radiuses are not small.

### Acknowledgements

The paper has benefited from discussions with Anthony R. Cassandra and Michael Littman. We also thank the associate editor Thomas L. Dean and the three anonymous reviewers for their insightful comments and suggestions and pointers to references. Research was supported by Hong Kong Research Council under grants HKUST 658/95E and Hong Kong University of Science and Technology under grant DAG96/97.EG01(RI).

## Appendix A: Proofs of Propositions 2 and 3

**Lemma 6** *Suppose $\mathcal{W}$ and $\mathcal{X}$ are two sets of state space functions. If $\mathcal{W}$ covers $\mathcal{X}$, then for any non-negative function $f(s)$,*

$$max_{\alpha \in \mathcal{W}} \sum_s \alpha(s) f(s) = max_{\beta \in \mathcal{X}} \sum_s \beta(s) f(s). \square$$

**Proof of Proposition 2:** Because of Proposition 1 and Lemma 2, it suffices to show that $\mathcal{Q}'_{a,o_+}$ covers $\mathcal{Q}_{a,o_+}$. By the definition of $\mathcal{Q}_{a,o_+}$ and Equation (10), we have, for any belief state $b$, that

$$
\begin{aligned}
\mathcal{Q}_{a,o_+}(b) &= max_{\alpha \in \mathcal{V}_{t-1}} \sum_s [\gamma \sum_{s_+} \alpha(s_+) P(s_+, o_+|s, a)] b(s) \\
&= \gamma \ max_{\alpha \in \mathcal{V}_{t-1}} \sum_{s_+} \alpha(s_+) [\sum_s b(s) P(s_+, o_+|s, a)]
\end{aligned}
$$

Since $\hat{\mathcal{V}}_{t-1}$ covers $\mathcal{V}_{t-1}$ and the term within the square brackets is a non-negative function of $s_+$, by Lemma 6 we have

$$
\begin{aligned}
\mathcal{Q}_{a,o_+}(b) &= \gamma \ max_{\alpha \in \hat{\mathcal{V}}_{t-1}} \sum_{s_+} \alpha(s_+) [\sum_s b(s) P(s_+, o_+|s, a)] \\
&= max_{\alpha \in \hat{\mathcal{V}}_{t-1}} \sum_s [\gamma \sum_{s_+} \alpha(s_+) P(s_+, o_+|s, a)] b(s) \\
&= \mathcal{Q}'_{a,o_+}(b),
\end{aligned}
$$

where the last equation is due to the definition of $\mathcal{Q}'_{a,o_+}$ and Equation (10). So, $\mathcal{Q}'_{a,o_+}$ does cover $\mathcal{Q}_{a,o_+}$. The proposition is proved. $\square$

**Lemma 7** *For any observation $z_+ = (o_+, R_+)$ of the region observable POMDP $\mathcal{M}'$,*

$$P(s_+, z_+|s, a) = 0,$$

*for any $s_+ \notin R_+$. $\square$*

Informally, this lemma says that the true state of the world must be in the region reported by the oracle.

**Lemma 8** *Let $\mathcal{W}$ and $\mathcal{X}$ be two sets of state space functions and $R$ be a region. If $\mathcal{W}$ covers $\mathcal{X}$ in region $R$, then for non-negative function $f(s)$ that is 0 when $s \notin R$, we have*

$$max_{\alpha \in \mathcal{W}} \sum_s \alpha(s) f(s) = max_{\beta \in \mathcal{X}} \sum_s \beta(s) f(s). \square$$

**Proof of Proposition 3**: Because of Proposition 1 and Lemma 4, it suffices to show that $\mathcal{Q}_{a,z_+,R}$ covers $\mathcal{Q}_{a,z_+}$ in region $R$, where $\mathcal{Q}_{a,z_+}$ is given by (11) with $o_+$ and $\mathcal{V}_{t-1}$ replaced by $z_+$ and $\mathcal{U}_{t-1}$.

Let $b$ be any belief state in $\mathcal{B}_R$. Similar to the proof of Theorem 2, we have

$$
\begin{aligned}
\mathcal{Q}_{a,z_+}(b) &= \gamma \; max_{\alpha \in \mathcal{U}_{t-1}} \sum_{s_+} \alpha(s_+)[\sum_s b(s)P(s_+, z_+|s, a)] \\
&= \gamma \; max_{\alpha \in \hat{\mathcal{U}}_{t-1,R_+}} \sum_{s_+} \alpha(s_+)[\sum_s b(s)P(s_+, z_+|s, a)] \\
&= max_{\alpha \in \hat{\mathcal{U}}_{t-1,R_+}} \sum_{s \in R}[\gamma \sum_{s_+} \alpha(s_+)P(s_+, z_+|s, a)]b(s) \\
&= max_{\beta \in \mathcal{Q}_{a,z_+,R}} \sum_s \beta(s)b(s) \\
&= \mathcal{Q}_{a,z_+,R}(b),
\end{aligned}
$$

where the second equation is true because of the fact that $\hat{\mathcal{U}}_{t-1,R_+}$ covers $\mathcal{U}_{t-1}$ in region $R_+$ and of Lemma 8. The term within the square brackets is a non-negative function of $s_+$ and it is 0 when $s_+ \notin R_+$ because of Lemma 7. The fourth equation is true because that $b(s){=}0$ when $s \notin R$. The proposition is proved. □

## Appendix B: Domination and Pruning

This appendix describes implementation of the procedures $\mathtt{dominate}(\alpha, \mathcal{W}, R, \epsilon)$, $\mathtt{purge}(\mathcal{W}, R)$, and $\mathtt{purge}(\mathcal{W})$. They were not given in the main text because they are minor adaptations of existing algorithms.

The procedure $\mathtt{dominate}(\alpha, \mathcal{W}, R, \epsilon)$ takes, as inputs, a state space function $\alpha$, a set of state space functions $\mathcal{W}$, a region $R$, and a nonnegative number $\epsilon$. It returns a belief state $b$ in $\mathcal{B}_R$ such that $\alpha(b){>}\mathcal{W}(b){+}\epsilon$. If such a belief state does not exist, it returns $\mathtt{nil}$. It can be implemented as follows.

Procedure $\mathtt{dominate}(\alpha, \mathcal{W}, R, \epsilon)$

- Inputs: $\alpha$ — A state space function,
  $\mathcal{W}$ — A set of state space functions,
  $R$ — A region, $\epsilon$ — A nonnegative number.

- Output: A belief state in $\mathcal{B}_R$ or $\mathtt{nil}$.

1. **If** $\mathcal{W}{=}\emptyset$, return an arbitrary belief state in $\mathcal{B}_R$.

2. Solve the following linear program:

   Variables: $x$, $b(s)$ for each $s \in R$.

   Maximize: $x$

   Constraints:

   $$
   \sum_{s \in R} \alpha(s)b(s) \geq x + \sum_{s \in R} \beta(s)b(s) \text{ for all } \beta \in \mathcal{W},
   $$

   $$
   \sum_{s \in R} b(s) = 1
   $$

   $$
   b(s) \geq 0 \text{ for all } s \in R.
   $$

3. **If** $x \leq \epsilon$, return **nil**, **else** return $b$.

The procedure **purge**$(\mathcal{W}, R)$ takes a set of state space functions $\mathcal{W}$ and a region $R$ and returns a set of state space functions that parsimoniously covers $\mathcal{W}$ in region $R$. To implement it, we need two subroutines.

A state space function $\alpha$ *pointwise dominates* another state space function $\beta$ in a region $R$ if $\alpha(s) \geq \beta(s)$ for all $s \in R$. The subroutine **pointwisePurge**$(\mathcal{W}, R)$ returns a minimal subset $\mathcal{W}'$ of $\mathcal{W}$ such that each state space function in $\mathcal{W}$ is pointwise dominated in the region $R$ by at least one state space function in $\mathcal{W}'$. Implementation of this subroutine is straightforward.

The subroutine **best**$(b, \mathcal{W}, R)$ returns a state space function $\alpha$ in $\mathcal{W}$ such that $\sum_{s \in R} b(s)\alpha(s) \geq \sum_{s \in R} b(s)\beta(s)$ for any other state space function $\beta$ in $\mathcal{W}$. Implementation of the subroutine is straightforward except for the issue of tie breaking. If the ties are not broken properly, **purge**$(\mathcal{W}, R)$ might return a regional covering of $\mathcal{W}$ that is not parsimonious. A correct way to break ties is as follows: Fix an ordering among states in $R$. This induces a lexicographic ordering among all state space functions. Among the tied state space functions, chose the one that is the largest under the lexicographic ordering (Littman, 1994).

The following implementation of **purge** is based on Lark's algorithm (White, 1991).

Procedure **purge**$(\mathcal{W}, R)$

- Inputs: $\mathcal{W}$ — A set of state space functions,
  $R$ — A region.

- Output: A set of state space functions that parsimoniously covers $\mathcal{W}$ in region $R$.

1. $\mathcal{W} \leftarrow$ **pointwisePurge**$(\mathcal{W}, R)$.
2. $\mathcal{X} \leftarrow \emptyset$.
3. **While** $\mathcal{W} \neq \emptyset$,
   (a) Pick a state space function $\alpha$ from $\mathcal{W}$.
   (b) $b \leftarrow$ **dominate**$(\alpha, \mathcal{X}, R, 0)$.
   (c) **If** $b =$ **nil**, remove $\alpha$ from $\mathcal{W}$.
   (d) **Else** remove **best**$(b, \mathcal{W}, R)$ from $\mathcal{W}$ and add it to $\mathcal{X}$.
4. Return $\mathcal{X}$.

Finally, the procedure **purge**$(\mathcal{W})$ takes a set of state space functions $\mathcal{W}$ and returns a parsimonious covering of $\mathcal{W}$. It can be implemented simply as follows.

Procedure **purge**$(\mathcal{W})$:

- **purge**$(\mathcal{W}, \mathcal{S})$.

Here, $\mathcal{S}$ is the set of all possible states of the world.